\documentclass[lettersize,journal]{IEEEtran}

\usepackage{amsmath,amsfonts}
\usepackage{algorithmic}
\usepackage{algorithm}

\usepackage{graphicx}
\usepackage{array}
\usepackage{booktabs}     
\usepackage{multirow}
\usepackage{stfloats}
\usepackage[caption=false,font=normalsize,labelfont=sf,textfont=sf]{subfig}

\usepackage[table]{xcolor} 
\usepackage[skins]{tcolorbox}

\definecolor{thermalred}{RGB}{220,50,47}
\definecolor{thermalblue}{RGB}{38,139,210}
\definecolor{thermalgray}{gray}{0.95}
\definecolor{mygreen}{RGB}{34,139,34}

\usepackage{pifont}        
\usepackage{textcomp}
\usepackage{url}
\usepackage{verbatim}
\usepackage{cite}
\usepackage{comment}

\newcommand{\cmark}{\ding{51}} 
\newcommand{\xmark}{\ding{55}}

\begin{document}

\title{Time-Reversed Imaging: Inferring Past Human–Environment Interactions from Multimodal Data}

\author{
Jorge Bacca, ~\IEEEmembership{IEEE Member},
Kebin Conteras,~\IEEEmembership{IEEE Student~Member},
Luis Toscano-Palomino,~\IEEEmembership{IEEE Student~Member},
Mauro Dalla Mura~\IEEEmembership{IEEE senior~Member}}



\maketitle

\begin{abstract}
We introduce time-reversed imaging, a new paradigm that infers what just happened in a scene from fading multimodal traces. Instead of extrapolating or interpolating video frames, our goal is to infer past human–environment interactions from residual physical imprints observable in thermal, ultraviolet, and visible spectra. To study this problem, we present TRACE-HEI, the first proof-of-concept dataset for time-reversed imaging, containing synchronized tri-modal video sequences of actions such as sitting, touching, moving objects, and liquid spills, captured across diverse materials and recorded up to three minutes after contact. To establish the benchmark, we proposed a multimodal inference approach that extracts structured textual descriptions of detected traces and uses them to constrain a vision-language-guided diffusion model for reconstructing plausible past frames. Experiments show that inferring recent events from fading traces is challenging but feasible when complementary modalities reduce solution ambiguity. This work defines the first computational and experimental foundation for time-reversed imaging, bridging vision, physics, and generative reasoning, opening new directions for scene understanding beyond instantaneous observation.
\end{abstract}


\section{Introduction}
\label{sec:intro}

\begin{figure}[!t]
    \centering
\includegraphics[width=\linewidth]{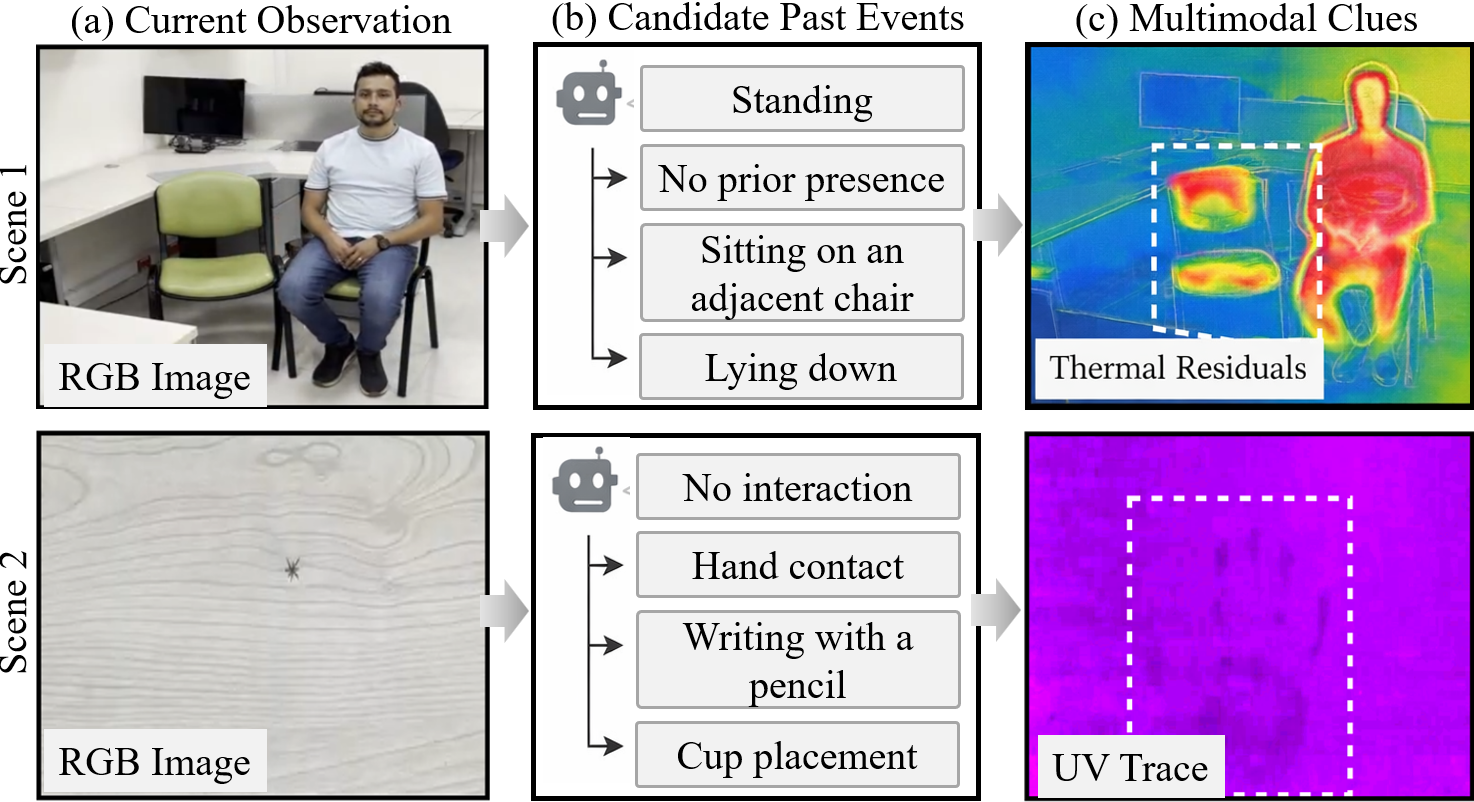}
\caption{\textbf{Ill-posedness of time-reversed inference and the role of physical traces.}
Given only a current RGB observation, the space of candidate past events is inherently underdetermined, as multiple interaction histories can lead to similar visual states. Multimodal clues captured through thermal and UV sensing reveal residual physical signatures of recent contact, providing additional constraints that make past-event inference significantly better conditioned.}
    \label{fig:motivation_image} 
\end{figure}

Inferring  \textit{what just happened} from post-event observations is an extremely ill-posed inverse problem. From a single RGB frame, the number of possible past events that could have produced the observed state is unbounded. As illustrated in Fig.~\ref{fig:motivation_image}, an RGB-only analysis cannot distinguish between fundamentally different scenarios, such as a person sitting on an adjacent chair versus a brief physical interaction with a surface.  This capability is critical in real-world  \textit{late-onset} sensing scenearios or when the coverage is incomplete. Investigators often arrive after an incident~\cite{ketsekioulafis2024artificial}, safety systems detect abnormal states without seeing the triggering action~\cite{koesdwiady2016recent}, and robots entering a workspace must infer recent human manipulations to act safely~\cite{robinson2023robotic}. Similarly, in smart homes or hospitals, systems observe outcomes, displaced objects, residual heat, or chemical traces, without recording the initiating event~\cite{cordoba2023human}. 

Traditional temporal reasoning methods such as video interpolation and extrapolation~\cite{gao2019disentangling,liu2017video,yu2019crevnet,lotter2016deep} operate exclusively in the forward direction of time, predicting future frames from past ones. Although these models demonstrate that temporal structure can be learned, and could, in theory, be reversed to infer plausible prior states, they rely entirely on statistical priors rather than on physical evidence. Consequently, inferring the past from RGB data alone is not just challenging; it is structurally underdetermined. Without additional clues, the space of possible preceding events remains inherently ambiguous.

However, real environments contain a form of physical memory. Human environment interactions leave behind residual traces that persist for seconds or even minutes after contact. These include thermal signatures, fluorescence changes, and subtle surface-level disturbances that are invisible to the human eye but detectable in the thermal (TH) and ultraviolet (UV) ranges. As shown in Fig.~\ref{fig:motivation_image}, TH imaging reveals heat left on chairs, while UV imaging exposes fluorescence variations caused by skin oils or sunscreen. These fading traces serve as measurable physical evidence of prior actions, substantially constraining the solution space of the inverse problem and rendering past-event inference far better conditioned than relying on RGB alone. Preliminary steps toward overcoming this ambiguity were recently explored in~\cite{contreras2025see}, where visual-language models (VLMs) were leveraged to perform time-reversed scene reconstruction from thermal traces alone across a limited set of images. However, that initial attempt was restricted to a single non-visible modality, evaluated on a small sample size, and lacked both a formal problem formulation and a standardized evaluation framework.

Extending our preliminary work~\cite{contreras2025see}, we introduce Time-Reversed Imaging, a new paradigm in computer vision that aims to infer recent past events using multimodal traces captured at the present moment. This paradigm seeks to recover the most plausible past scene state by exploiting the physical decay dynamics of UV and thermal residual signals. As an emerging research direction, this paradigm requires the establishment of foundational elements: (i) a formal problem definition, (ii) dedicated multimodal datasets capturing residual trace dynamics, (iii) principled inference methods grounded in physical evidence, and (iv) a standardized benchmark for evaluating reverse-time reasoning.

To support this new task, we introduce TRACE-HEI, a proof-of-concept multimodal dataset specifically designed for time-reversed scene understanding. TRACE-HEI includes 100 synchronized  UV + RGB + TH video sequences of controlled human–environment interactions such as sitting, touching, moving objects, handling materials (wood, fabric, metal, plastic, glass) and spreading or cleaning liquids (water, oil, sunscreen). Each action occurs within the first 30s, and sequences are then recorded for the following 3 minutes, capturing the temporal decay of heat, fluorescence, and surface disturbance signals. The resulting multi-modal, multi-material, and multi-action dataset provides the first foundation for studying how physical traces evolve over time and how they constrain inference of the recent past.

Furthermore, we present a training-free inference framework that first extracts structural information from present-time frames (UV + RGB + TH) by predicting both the preceding event category and the involved object. These predictions act as a structured semantic description, dramatically reducing the ambiguity of possible past states. To transform these semantic predictions into an image, we couple them with a VLM-guided diffusion process, in which the predicted event and material are encoded as textual prompts. This generative process is thus anchored to the multimodal physical traces and conditioned on high-level semantic evidence, ensuring images that are physically consistent, semantically plausible, and faithful to the scene’s geometry and illumination. 

Since no benchmark previously exists for this problem, we further establish the first unified evaluation protocol for time-reversed scene inference. The proposed benchmark jointly evaluates semantic event prediction and visual reconstruction through complementary low-level, perceptual, and high-level semantic metrics, enabling systematic and reproducible comparison across methods.  Our experiments demonstrate that multimodal fusion significantly outperforms single-modality approaches. Thermal and UV traces provide complementary information, and their combination with RGB-VIS yields robust generalization to unseen subjects and materials. Furthermore, by incorporating structural and semantic constraints, our model generates reconstructions that remain faithful to both the physical evidence and the scene context. This approach establishes a new frontier for scene understanding beyond the visible present, toward recovering the recent past. To summarize, this work makes the following novel contributions:
\begin{itemize}
    \item \textbf{Problem Formulation:} We formalize time-reversed imaging as a new computational task, inferring the recent past from present-time multimodal residual signals, extending our preliminary work~\cite{contreras2025see} to a principled theoretical and mathematical framework.
    \item \textbf{Dataset Creation:} We substantially extend the preliminary dataset introduced in our earlier proof-of-concept study~\cite{contreras2025see}. Specifically, we expand the benchmark to 100 recorded scenes and introduce UV sensing as an additional modality alongside TH and RGB imaging. The resulting dataset, named TRACE-HEI, constitutes the first multimodal benchmark explicitly designed to evaluate past scene reconstruction from residual traces across diverse actions, materials, and interaction scenarios.
    \item \textbf{Inference Method:} We develop a multimodal past-event inference framework that integrates semantic reasoning with VLM-guided diffusion, bridging physical evidence and semantic priors.
    \item \textbf{Benchmark Protocol:}  We propose a comprehensive evaluation protocol spanning low-level reconstruction metrics, feature-level consistency measures, and high-level semantic assessments for systematic comparison.
\end{itemize}

\section{Related Work}

\textbf{Forward Temporal Prediction and Video Reasoning.} A large body of work focuses on forecasting future visual states from past
observations, including video interpolation, extrapolation, and motion
prediction~\cite{liu2017video,gao2019disentangling,lotter2016deep,yu2019crevnet}.
These models learn temporal dynamics by operating in the forward direction of
time, reconstructing unseen frames based on previously observed content.
While they demonstrate strong predictive capability, they rely primarily on
statistical regularities rather than physical evidence, and thus cannot
reliably infer prior events from single post-interaction observations.

\textbf{Multimodal Imaging Beyond the Visible Spectrum.}
Thermal and ultraviolet imaging have enabled progress in
contact detection, material analysis, and forensics, where heat transfer and
fluorescence reveal otherwise imperceptible signatures~\cite{patel2019potential,cardone2017new}.
Thermal traces have been used to estimate recency of contact with objects or
surfaces, but mainly in controlled, single-object settings~\cite{brahmbhatt2019contactdb}.
UV fluorescence has similarly been used to detect skin oils, residues, and
surface contamination~\cite{de2020possible}, yet rarely in a combined, temporally aligned
multimodal setting. To date, no dataset systematically captures RGB, UV, and TH modalities after human–environment interactions across diverse materials, actions, and trace-decay intervals.

\textbf{Multimodal Clues for Post-Event Reconstruction.}
Complementary sensing modalities provide measurable physical evidence that
reduces the ambiguity inherent to recovering past events. Event cameras
capture microsecond-level brightness changes for motion estimation and
video reconstruction~\cite{zhu2024video,pan2020high}, but their streams
encode only ongoing illumination changes; once motion stops, they contain no
information about prior interactions. Thermal imaging offers a different
form of physical memory by capturing long-wave infrared radiation that
persists after contact. Tang et al.~\cite{tang2023happened} showed that
fading heat patterns can recover human poses a few seconds after
interaction, but their approach is restricted to pose estimation, short
temporal windows, and a single modality. Our preliminary results in~\cite{contreras2025see}
extended this idea by combining RGB and thermal cues with a diffusion model. Nevertheless, that exploration relied on only a few curated examples and lacked systematic data collection, multimodal sensing, or large-scale evaluation. To the best of our knowledge, the task of \emph{Time-Reversed Imaging} that we introduce here has not been formally defined, benchmarked, or explored as a general vision problem.

\begin{figure*}[!t]
    \centering
    \includegraphics[width=0.85\linewidth]{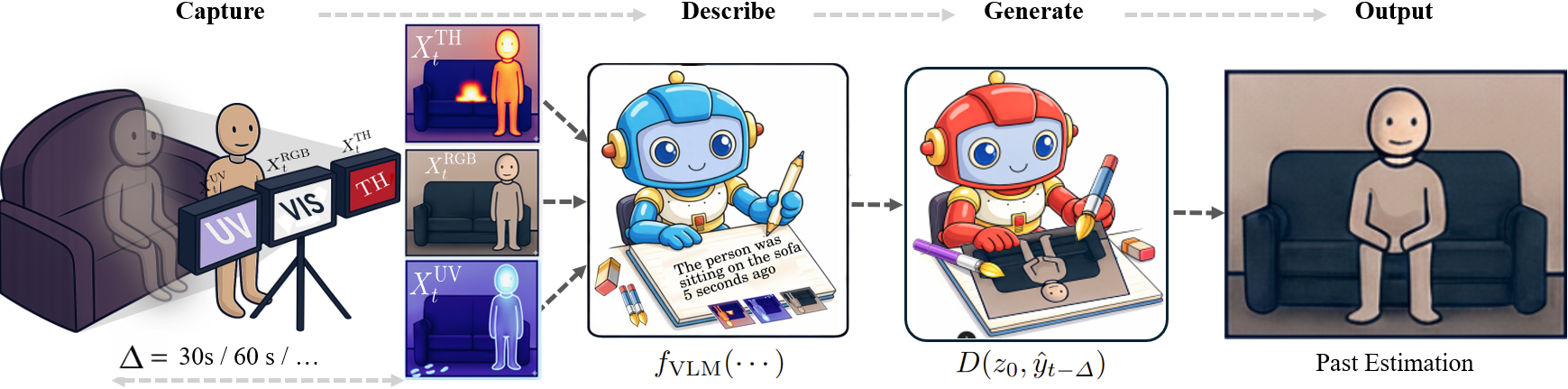}

\caption{
\textbf{Time-reversed inference framework.}
Synchronized RGB, UV, and TH observations captured at the present time
provide multimodal physical evidence of past interactions. A VLM predicts the most likely preceding event, and other VLM-guided
diffusion model infers the image as it appeared at time $t-\Delta $.
}
    \label{fig:overview_framework}
\end{figure*}
\section{Time-Reversed Imaging}

The goal of Time-Reversed Imaging is to infer the most plausible past scene state, that is, to describe and reconstruct what just happened, based on multimodal measurements captured after the event. Figure~\ref{fig:overview_framework} provides an overview of the proposed framework, which integrates (i) multimodal sensing, (ii) structured semantic prompts, and (iii) VLM-guided constrained diffusion for visual reconstruction. This framework is fundamentally model-agnostic, defining time-reversed imaging as a general conditional inverse inference task, independent of specific architectural configurations.

\subsection{Problem Formulation}
Let $x_t^{RGB} \in \mathbb{R}^{H \times W \times 3}$, 
$x_t^{UV} \in \mathbb{R}^{H \times W}$, and 
$x_t^{TH} \in \mathbb{R}^{H \times W}$ 
denote the synchronized RGB (visible), UV, and thermal images captured at the current time $t$, after a human--environment interaction 
has occurred and $\Delta>0$ as the temporal offset of interest.
We divided the problem into two parts:
(1) to infer the most probable past interaction label 
$y_{t-\Delta}$, describing which action occurred and which objects were involved, for example, \textit{“a person was sitting on the chair”}; and (2) to reconstruct a plausible RGB image 
$x_{t-\Delta}^{RGB}$ showing how the scene appeared $\Delta$ seconds earlier. We formalize this inverse temporal reasoning task as estimating the conditional probability distribution
\begin{equation}
p(x_{t-\Delta}^{RGB}, y_{t-\Delta} \mid \mathcal{E}_t, \Delta),
\label{eq:jointposterior}
\end{equation}
which represents the posterior over possible past scene states, given the 
present multimodal evidence $\mathcal{E}_t =\{x_t^{RGB}, x_t^{UV}, x_t^{TH}\}$ conditioning on 
$\Delta$.

Direct estimation of the joint posterior in Eq.~\ref{eq:jointposterior} 
is impractical. Consequently, we decompose the distribution 
using the chain rule of probability, separating the estimation of the 
past event $y_{t-\Delta}$  from the reconstruction of its visual 
appearance $x_{t-\Delta}^{RGB}$
as
\begin{equation}
p(x_{t-\Delta}^{RGB}, y_{t-\Delta} \mid \mathcal{E}_t, \Delta)
= p(x_{t-\Delta}^{RGB} \mid \mathcal{E}_t, y_{t-\Delta}, \Delta)        p(y_{t-\Delta} \mid \mathcal{E}_t, \Delta).
\label{eq:factorization}
\end{equation}

This factorization enables a two-stage inference procedure in which 
a classifier first infers the most probable past interaction, followed by 
a conditional generative model that reconstructs the RGB appearance of the 
scene consistent with both the predicted semantics and the observed traces.

\subsection{Structured Past-Event Description}

Due to the limited multimodal datasets for past-event reasoning, we adopt a training-free strategy to infer a semantic hypothesis conditioned on the multimodal observations $(x_t^{RGB}, x_t^{TH}, x_t^{UV})$. Rather than generating unconstrained free-form text, we introduce a \emph{Structured Past-Event Description (SPED)}, which restricts the output space and provides stable conditioning for downstream generative reconstruction.

The core idea is to leverage a VLM to interpret residual physical traces across modalities and translate them into a compact, structured semantic representation of the recent interaction. This representation acts as an intermediate abstraction layer, linking physical evidence to visual reconstruction. Formally, we estimate
\begin{equation}
\hat{y}_{t-\Delta} = f_{\mathrm{VLM}}(x_t^{RGB}, x_t^{TH}, x_t^{UV}, \Delta, \text{SPED}),
\end{equation}
where $\hat{y}_{t-\Delta}$ denotes a constrained semantic hypothesis describing the most plausible interaction occurring $\Delta$ seconds earlier, and the SPED is proposed as

\begin{tcolorbox}[
  enhanced,
  colback=thermalgray,
  colframe=thermalred!20!thermalblue,
  coltitle=black,
  fonttitle=\bfseries,
  title=SPED: Structured Past-Event Description,
  boxrule=0.2pt,
  arc=2mm,
  left=1mm, right=1mm, top=1mm, bottom=1mm
]
\small\texttt{\textcolor{thermalred!20!thermalblue}{Using the provided thermal and UV traces, infer the event that most likely occurred 
$\langle \Delta \rangle$ seconds earlier}. Strictly follow this structure:
The most recent event is that $\langle object \rangle$ was $\langle action \rangle$ 
[optionally involving $\langle object_2 \rangle$ or $\langle action_2 \rangle$].}
\end{tcolorbox}

This structured prompting constrains semantic variability, reducing ambiguity while encouraging physically consistent interpretations of the observed thermal and UV decay patterns. In this work, we intentionally restrict the set of possible interactions to relatively simple scenarios in order to isolate and demonstrate the feasibility of the past-image reconstruction problem, which remains significantly more challenging than the semantic event prediction itself (See the Supplementary Material (SM) for a detailed analysis). 

\subsection{VLM-Guided Diffusion for Past-Scene Reconstruction}

Given the estimated structured past event $\hat{y}_{t-\Delta}$, we reconstruct the RGB
appearance of the scene at time $t-\Delta$ by sampling from the conditional
distribution
\begin{equation}
p(x_{t-\Delta}^{RGB} \mid \mathcal{E}_t, \hat{y}_{t-\Delta}),
\end{equation}
which we approximate using a diffusion model guided by a pretrained VLM. Starting from a Gaussian latent 
$z_T \sim \mathcal{N}(0, I)$, the denoising process iteratively applies
\begin{equation}
z_{t-1} = g_\phi(z_t, \mathcal{E}_t, \hat{y}_{t-\Delta}),
\end{equation}
until the final latent is decoded into an RGB estimate of the past scene:
\begin{equation}
\hat{x}_{t-\Delta}^{RGB} = D(z_0, \hat{y}_{t-\Delta}).
\end{equation}
The VLM embeds the $\hat{y}_{t-\Delta}$ using a \textit{Reconstruction Prompt}, described below, and injects its semantic
representation into the diffusion model via cross-attention, constraining the generation process to remain physically consistent with thermal/UV traces while adhering to the inferred event semantics~\cite{google2025gemini2_5}. The final reconstruction $\hat{x}_{t-\Delta}^{RGB}$ thus constitutes a
sample from the conditional posterior, integrating both multimodal evidence and high-level semantic cues.

\begin{tcolorbox}[
  enhanced,
  colback=thermalgray,
  colframe=thermalred!20!thermalblue,
  coltitle=black,
  fonttitle=\bfseries,
  title=Reconstruction Prompt,
  boxrule=0.2pt,
  arc=1mm,
  left=1mm, right=1mm, top=1mm, bottom=1mm
]
\small\texttt{\textcolor{thermalred!20!thermalblue}{
Edit the RGB image to reconstruct the scene as it appeared $<$$\Delta$$>$ seconds earlier, preserving the environment, 
lighting, viewpoint, and color distribution.} Assume that the recent event was that 
$<$$\hat{y}_{t-\Delta}$$>$
as indicated by the current thermal and UV traces.}
\end{tcolorbox}

\section{Experimental Protocol}

Since no prior benchmark exists for time-reversed scene inference, we first establish a unified evaluation protocol that jointly assesses semantic event prediction and visual reconstruction quality. This benchmark combines low-level image fidelity, perceptual similarity, semantic consistency, and object-level localization metrics, providing a comprehensive assessment of reverse-time inference.

\subsection{TRACE-HEI Dataset}

\begin{figure}[!t]
    \centering
        \includegraphics[width=\linewidth]{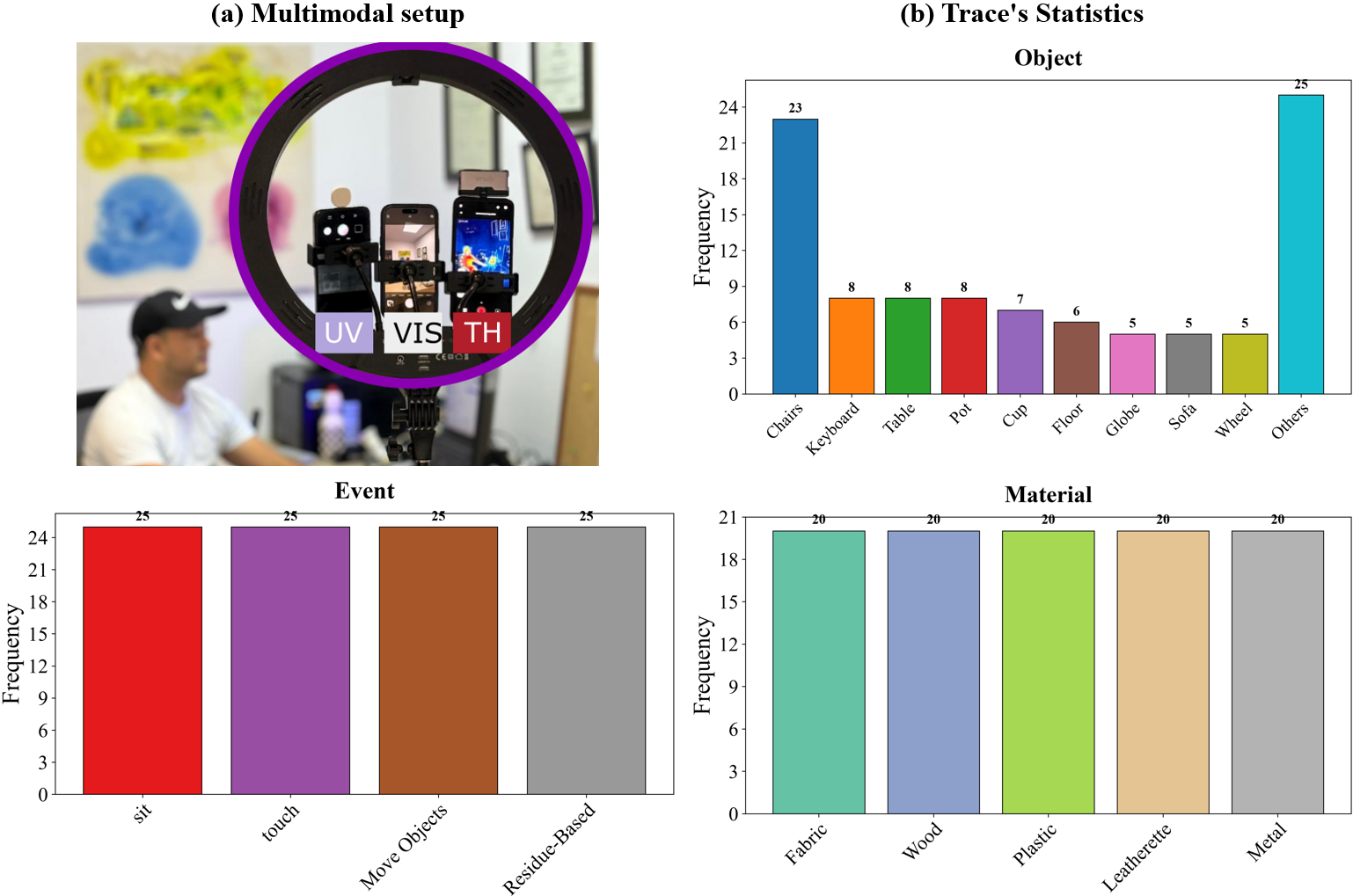}
    \caption{\textbf{TRACE-HEI summary.} (\textbf{a}) Multimodal acquisition setup with UV, RGB, and TH sensors synchronized to capture post-interaction residual traces. (\textbf{b}) Statistics of the 100-scene subset used for inference (actions, objects, gender, and materials).}
    \label{fig:tracehei_setup_all_es}
\end{figure}

We introduce\footnote{https://www.kaggle.com/datasets/pigroup/trace-hei} the first multimodal dataset comprising synchronized RGB (visible), thermal (TH), and ultraviolet (UV) sequences recorded during and after controlled human–environment interactions. Data acquisition was performed with three sensors: (i) an iPhone 15 RGB camera (\(4032\times3024\), visible band 400--700\,nm), (ii) a FLIR ONE Pro thermal camera (\(160\times120\), LWIR 8--14\(\,\mu\)m), and (iii) a 365\,nm UV camera with dedicated UV-A illumination (\(1920\times1080\), response in 320--400\,nm). The physical setup and sensor placement are illustrated in Figure~\ref{fig:tracehei_setup_all_es}\textbf{(a)}. All scene videos are processed with a unified pipeline for temporal synchronization, geometric correction, and multimodal normalization. Specifically, we correct spatial misalignments across sensors, apply positional correction to ensure pixel-level correspondence across modalities, and, due to the UV sensor optical configuration, perform horizontal inversion (\textit{flip}) of UV captures when required to maintain orientation consistency with RGB/TH. In addition, all modalities are normalized per sequence to reduce illumination changes, thermal drift, and dynamic-range differences across sensors. Additional hardware and calibration details are provided in the SM.

The acquisition protocol consists of two phases: (i) \textit{Interaction Phase (0–30s),} A controlled human–environment interaction is performed and maintained for up to 30 seconds. This phase defines the ground-truth past event and the corresponding scene state that the proposed framework aims to reconstruct.

(ii) \textit{Decay Phase (30-180s).} After the interaction concludes, we record a 180-second multimodal sequence to analyze the temporal evolution and dissipation of physical traces. This phase enables the study of how long modality-specific cues persist and up to what delay $\Delta$ reliable past-scene inference remains feasible.

The dataset comprises 100 multimodal scenes spanning four main interaction categories: sit/stand transitions, touch interactions, object manipulation, and residue-based interactions. As shown in Figure~\ref{fig:tracehei_setup_all_es}\textbf{(b)}, the action distribution is balanced, with uniform representation across all four categories (25 scenes each), ensuring unbiased evaluation of temporal inference performance. The object distribution reflects realistic indoor environments, including chairs, keyboards, tables, cups, floor, globes, sofa, wheel, and other objects. This diversity introduces variability in geometry, contact area, and thermal capacity, which directly influences trace formation and decay dynamics. Material composition further enriches the dataset, encompassing fabric, wood, plastic, leatherette, and metal. These materials exhibit distinct thermal conductivity, emissivity, absorption, and fluorescence properties, creating heterogeneous physical trace behaviors across modalities. Specifically, thermal imagery captures short to mid-term heat transfer and dissipation patterns governed by material-specific diffusion processes, while UV imagery reveals fluorescence and residue cues whose persistence depends on surface texture and substance composition. Visual examples of the dataset are provided in the SM.

\subsection{Benchmark and Evaluation Metrics}

Consistent with the probabilistic formulation in Eq.~\eqref{eq:factorization}, the proposed benchmark independently evaluates the two complementary outputs of the time-reversed imaging framework: (i) semantic past-event inference and (ii) past-frame reconstruction. Since these tasks capture fundamentally different aspects of reverse-time inference, we employ dedicated evaluation metrics for each one.
\subsubsection{Past-Event Inference}

Since Vision-Language Models (VLMs) generate free-form textual outputs, each prediction is converted into a structured representation composed of semantic attributes describing the inferred interaction. Specifically, every prediction is decomposed into three semantic components: (i) the \textit{action}, (ii) the \textit{interacting object}, and (iii) the \textit{material} of the interacting object. Based on these structured annotations, we report the following evaluation metrics:

\paragraph{Action Accuracy}
Percentage of samples for which the predicted action matches the ground-truth interaction.

\paragraph{Object Accuracy}
Percentage of samples for which the predicted interacting object matches the ground-truth object.

\paragraph{Material Accuracy}
Percentage of samples for which the predicted material matches the annotated material category.

\paragraph{Exact Match}
Percentage of samples for which the action, object, and material are all correctly predicted simultaneously.

\paragraph{Top-3 Accuracy}
Percentage of samples for which the correct interaction appears among the top three candidate hypotheses generated by the VLM. This metric is particularly relevant given the inherent ambiguity of inferring past events from post-event observations.

\subsubsection{Past-Frame Reconstruction}

Reconstructing a past frame requires recovering not only low-level image details but also semantically consistent scene content. To comprehensively evaluate reconstruction quality, we adopt a hierarchical benchmark encompassing low-level fidelity, feature-level perceptual similarity, and high-level semantic consistency.

\paragraph{Low-Level Metrics}

We report the Peak Signal-to-Noise Ratio (PSNR)~\cite{wang2004image} and the Structural Similarity Index Measure (SSIM)~\cite{wang2004image} to quantify pixel-level reconstruction fidelity. PSNR measures the absolute intensity differences between the reconstructed image $\hat{x}_{t-\Delta}^{RGB}$ and the ground-truth frame $x_{t-\Delta}^{RGB}$, while SSIM evaluates the preservation of local structural information.

\paragraph{Feature-Level Metrics}

To evaluate perceptual similarity beyond strict pixel-wise correspondence, we employ the Learned Perceptual Image Patch Similarity (LPIPS) metric~\cite{zhang2018unreasonable}, which measures distances between deep feature representations of $\hat{x}_{t-\Delta}^{RGB}$ and $x_{t-\Delta}^{RGB}$. In addition, we compute the cosine similarity between CLIP visual embeddings~\cite{radford2021learning} extracted from the reconstructed and ground-truth images. This embedding-based metric assesses global semantic consistency, providing a complementary evaluation of whether the language-guided diffusion process preserves the semantic content of the original scene.

\paragraph{High-Level Metrics}

To assess semantic and spatial consistency at the object level, we report two complementary metrics. First, we compute the Overall Accuracy (OA)~\cite{taha2015metrics}. Specifically, object masks are obtained using the Segment Anything Model (SAM)~\cite{kirillov2023segment} for both the reconstructed image and the ground-truth frame. OA is then computed as the pixel-wise agreement between the corresponding object regions, measuring whether objects are correctly recovered and spatially aligned without emphasizing precise boundary delineation. Second, we evaluate object localization using a pretrained YOLOv11 detector~\cite{jocher2024ultralytics}. The primary interacting objects are detected in both reconstructed and reference images, and the Intersection-over-Union (IoU)~\cite{everingham2010pascal} is computed between the corresponding bounding boxes. While OA captures overall semantic consistency, IoU specifically measures the accuracy of the reconstructed object locations and their spatial overlap with the ground truth.

\section{Experimental Results}

Using the experimental protocol introduced in the previous section, we evaluate the proposed time-reversed imaging framework through a series of complementary experiments. We first assess the semantic reasoning capabilities of different Vision-Language Models (VLMs) for past-event inference. Next, we analyze the impact of structured prompting, multimodal sensing, and the choice of VLM-guided generator on past-frame reconstruction. Finally, we investigate the temporal limits of the proposed framework and its robustness to physics-informed estimation of the elapsed time. All experiments are conducted on the TRACE-HEI dataset using the fully training-free framework described in Section~3, and results are reported as the average over five independent runs.

\subsection{VLM Comparison for Past-Event Description}

We first evaluate how different VLMs, including GPT-5~\cite{singh2025openai}, Gemini-3 Pro~\cite{team2023gemini}, Claude~3.5~\cite{anthropic_claude3}, Qwen~3.5-Plus~\cite{wang2024qwen2}, and LLaVA-NeXT~\cite{liu2024improved} perform when generating structured past-event descriptions from multimodal observations. In this analysis, we focus on the case $\Delta=30$ seconds. At this temporal offset, thermal and UV signals still provide meaningful physical evidence of recent interactions while the event itself is no longer visible in the RGB observation.

\begin{table}[b]
\centering
\caption{Evaluation of structured past-event descriptions across different VLMs.}
\label{tab:vlm_semantic}
\setlength{\tabcolsep}{1pt} 
\footnotesize 
\begin{tabular}{lccccc}
\toprule
VLM Model & Action ($\uparrow$) & Object ($\uparrow$) & Material ($\uparrow$) & Exact Match ($\uparrow$) & Top-3 ($\uparrow$) \\
\midrule
GPT-5        & \textbf{88.4} & \underline{91.2} & \textbf{85.7} & \textbf{78.3} & \textbf{94.6} \\
Gemini 3 Pro & \underline{86.9} & \textbf{91.5} & \underline{84.2} & \underline{76.1} & \underline{93.8} \\
Claude 3.5   & 84.2          & 87.8          & 82.1          & 72.5          & 91.2          \\
Qwen3.5 Plus & 79.5          & 82.3          & 76.4          & 65.8          & 86.4          \\
LLaVA-NeXT   & 72.1          & 75.4          & 68.9          & 58.2          & 79.1          \\
\bottomrule
\end{tabular}
\end{table}

Results in Table ~\ref{tab:vlm_semantic} show a performance hierarchy led by GPT-5 and Gemini 3 Pro, which both demonstrate superior multispectral reasoning. GPT-5 achieves top marks across all metrics, notably reaching 88.4\% Action Accuracy and a 94.6\% Top-3 Accuracy. While Claude 3.5 remains highly precise in object identification, it slightly trails the leaders in Material Accuracy (82.1\%). A significant performance gap appears with Qwen3.5 Plus and LLaVA-NeXT, particularly in the material and exact match categories; these models struggle to map residual thermal signatures to specific physical properties, whereas frontier-class models successfully bridge the gap between abstract heat traces and structured semantic descriptions.

\subsection{Prompt Structure Analysis and Robustness}
To address the sensitivity of the reconstruction to linguistic variations, we conduct a comparative analysis of different prompting strategies. This study evaluates how the density and structure of the semantic prior influence the generative model's ability to recover the past state. We contrast three distinct configurations: (i) Free-form Description, which utilizes only the initial task instruction without structural constraints; (ii) Reasoning-based Prompt, which encourages the VLM to perform a "chain-of-thought" explanation of the thermal/UV traces before concluding; and (iii) the proposed Structured Past-Event Description (SPED), which enforces a strict semantic template focused on interaction slots. As shown in Table~\ref{tab:prompt_robustness}, providing raw descriptions (Free-form) leads to lower low-level fidelity (PSNR: 17.05). While the VLM may correctly identify the event, the lack of structural focus introduces linguistic "noise" that can misguide the diffusion prior. The reasoning-based approach improves results by ensuring physical consistency, but our SPED strategy yields the best performance in High-Level metrics (AO: 0.854, IoU: 0.739). This confirms that by constraining the output to specific semantic slots, we minimize hallucination and maximize the alignment between the inferred physical interaction and the visual reconstruction engine. 
\begin{table}[t]
\centering 
\caption{\small Comparative analysis of prompt structures for past-event reconstruction at $\Delta = 30$s.}
\vspace{-0.7em}
\label{tab:prompt_robustness}
\renewcommand{\arraystretch}{1.3}
\resizebox{\columnwidth}{!}{
\begin{tabular}{l|cc|cc|cc}
\toprule
\textbf{Prompting Strategy}
& \multicolumn{2}{c|}{\textbf{Low-Level}}
& \multicolumn{2}{c|}{\textbf{Feature-Level}}
& \multicolumn{2}{c}{\textbf{High-Level}} \\
\cmidrule(lr){2-3} \cmidrule(lr){4-5} \cmidrule(lr){6-7}
& \textbf{PSNR $\uparrow$} & \textbf{SSIM $\uparrow$}
& \textbf{LPIPS $\downarrow$} & \textbf{CLIP $\uparrow$}
& \textbf{AO $\uparrow$} & \textbf{IoU $\uparrow$} \\
\midrule
Free-form description 
& 17.056 & 0.682 
& 0.320  & \textbf{93.938} 
& 0.765  & 0.624  \\

Reasoning (CoT) prompt 
& \underline{17.842 } & \underline{0.701 } 
& \underline{0.308 } & 91.820 
& \underline{0.812} & \underline{0.698 } \\

\textbf{SPED (Proposed)} 
& \textbf{18.268 } & \textbf{0.718 } 
& \textbf{0.297 } & \underline{92.755 } 
& \textbf{0.854 } & \textbf{0.739} \\
\bottomrule
\end{tabular}
}
\vspace{-1em}
\end{table}

\begin{figure}[!t]
    \centering \vspace{-1.5em}
\includegraphics[width=1\linewidth]{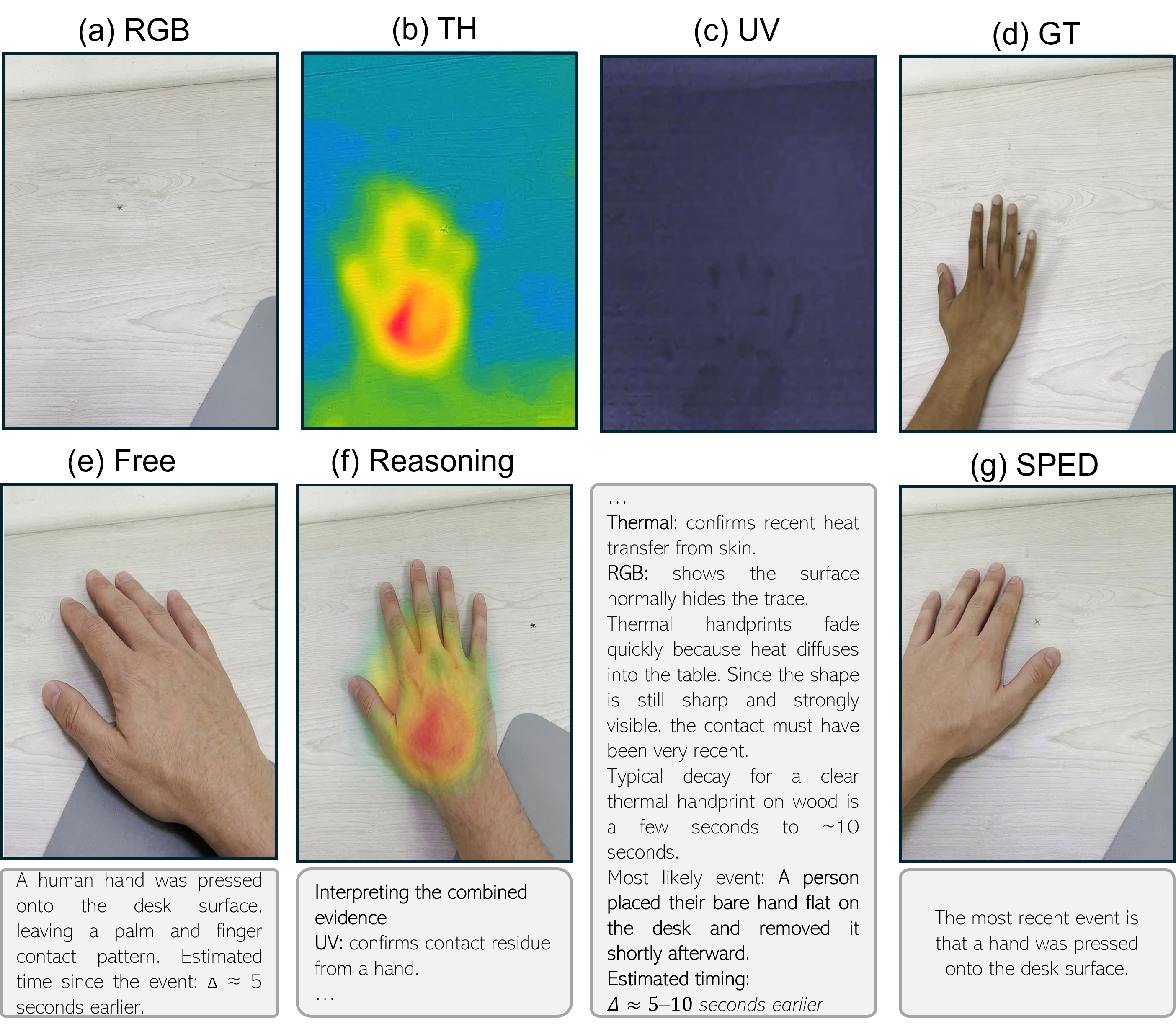} \vspace{-1.5em}
    \caption{A visual example of the prompt structure analysis. (a-c) The RGB, TH, and UV inputs, and (d) is the ground truth. (e-g) Shows the semantic results and the corresponding reconstructions for the Free-form, Reasoning, and SPED strategies.} 
\label{fig:prompt_robustness}
\end{figure}

The visual evidence in Figure \ref{fig:prompt_robustness} further supports the quantitative findings that prompt structure is a critical determinant of reconstruction fidelity. While the Free-form description (e) successfully identifies the "hand press" event, the resulting image displays a lack of morphological precision and skin-tone consistency, explaining the lower PSNR/SSIM scores in Table~\ref{tab:prompt_robustness}. Interestingly, the Reasoning approach (f), though physically accurate in its "chain-of-thought" deduction, occasionally causes the diffusion model to misinterpret the description of heat traces as a request for visual heatmaps, leading to the thermal artifacts seen in the output. In contrast, our SPED strategy (g) eliminates linguistic ambiguity by filtering the VLM's inference into a streamlined, interaction-focused prompt. This ensures that the generative engine receives only the necessary spatial and semantic constraints required to recover the past state, resulting in a reconstruction that is both perceptually natural and physically grounded.

\subsection{Impact of Semantic Descriptions on Reconstruction}

To quantify the dependence of visual fidelity on semantic accuracy, we evaluate the reconstruction pipeline by varying the VLM-generated structured prompt while keeping the diffusion engine (Gemini 3) fixed, using SPED. This controlled setting isolates the influence of the initial semantic inference on the final image synthesis. Experiments were conducted at a temporal offset of $\Delta = 60$ s, a challenging regime where visual cues are absent, and the model must rely entirely on the physical consistency of the VLM's predictions.

\begin{table}[b]
\centering
\caption{Influence of Semantic Descriptions on Time-Reversed Visual Reconstruction (60\,s Delay). }
\label{tab:vlm_evaluation}
\setlength{\tabcolsep}{4.0pt}
\renewcommand{\arraystretch}{1.2}
\resizebox{\columnwidth}{!}{
\begin{tabular}{l|cc|cc|cc}
\toprule
\textbf{Method}
& \multicolumn{2}{c|}{\textbf{Low-Level}}
& \multicolumn{2}{c|}{\textbf{Feature-Level}}
& \multicolumn{2}{c}{\textbf{High-Level}} \\
\cmidrule(lr){2-3}\cmidrule(lr){4-5}\cmidrule(lr){6-7}
& PSNR $\uparrow$ & SSIM $\uparrow$
& LPIPS $\downarrow$ & CLIP $\uparrow$
& AO $\uparrow$ & IoU $\uparrow$ \\
\midrule
GPT-5
& 16.01
& 0.615
& \textbf{0.335}
& \textbf{92.67}
& \textbf{0.912}
& \textbf{0.851} \\

Gemini 3 Pro
& \textbf{16.07}
& \textbf{0.622}
& \underline{0.338}
& \underline{92.54}
& \underline{0.908}
& \underline{0.848} \\

Claude 3.5
& \underline{16.02}
& \underline{0.618}
& 0.342
& 92.41
& 0.902
& 0.842 \\

Qwen3.5 Plus
& 14.12
& 0.485
& 0.452
& 85.33
& 0.784
& 0.712 \\

LLaVA-NeXT
& 13.45
& 0.421
& 0.518
& 81.20
& 0.695
& 0.623 \\
\bottomrule
\end{tabular}}
\end{table}

\begin{figure}[t]
    \centering\includegraphics[width=\linewidth]{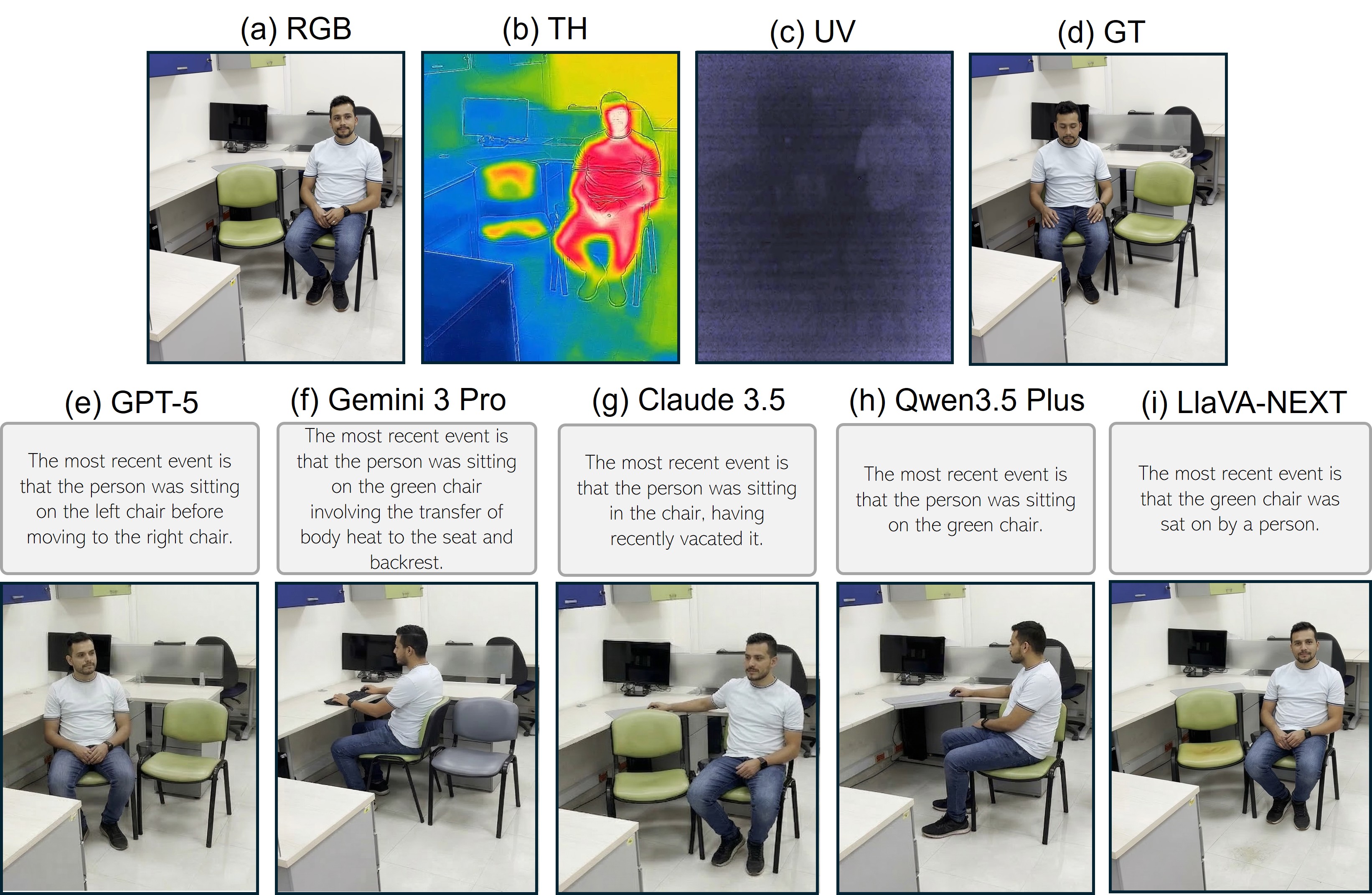}
    \caption{Comparative analysis of reconstruction fidelity based on VLM semantic inference at $\Delta = 60$ s. (a-c) Multimodal inputs including RGB, Thermal (TH), and Ultraviolet (UV) channels. (d) Ground Truth (GT) state $\Delta$ seconds prior. (e-i) Reconstruction results using different VLM-generated semantic prompts (top row) to guide the diffusion engine.}
    \label{fig:examples_section2_2}
\end{figure}

The results in Table~\ref{tab:vlm_evaluation} demonstrate that the visual reconstruction quality is highly robust when leveraging frontier-class models as semantic proxies. GPT-5, Gemini 3 Pro, and Claude 3.5 exhibit nearly identical performance across all tiers, with high-level alignment scores exceeding 0.91 (AO) and 0.84 (IoU). This suggests that these models have reached a stable level of reliability in decoding residual thermal-UV traces and converting them into structured signals. In contrast, Qwen3.5 Plus and LLaVA-NeXT show a significant performance degradation, particularly in the High-Level and Feature-Level metrics. The higher LPIPS and lower IoU scores for these models indicate that inaccuracies in the semantic inference stage—such as misidentifying materials or object boundaries—propagate through the pipeline, leading to reconstructions that are both perceptually inconsistent and physically inaccurate relative to the ground-truth event.

Figure~\ref{fig:examples_section2_2} shows a visual example of this experiment. It can be seen that GPT-5 and Gemini 3 Pro (e, f) provide precise spatial descriptions that allow the model to accurately relocate the subject to the left chair, closely matching the Ground Truth (d). In contrast, models like LLaVA-NeXT (i) generate generalized or incomplete descriptions—such as "the green chair was sat on by a person"—without specifying the necessary temporal or positional shifts. This results in the subject remaining in the "current" position from the input RGB image, failing to achieve a true time-reversed reconstruction. These failures underscore that for high-fidelity temporal back-projection, the bottleneck lies not in the generative capacity of the diffusion model, but in the semantic depth of the initial multimodal inference.

\subsection{Ablation Study}

To analyze the contribution of each component in the proposed framework, we conduct a systematic ablation study focusing on the impact of each sensing modality (thermal and UV) and the role of structured past-event descriptions in constraining the generative process. For controlled comparison, we fix the reconstruction horizon to $\Delta = 30$ seconds. For the past-event description, we employ ChatGPT-5.2~\cite{openai2023dalle3} as the VLM. For image generation, we use Gemini 2.5~\cite{google2025gemini2_5} as the VLM-guided diffusion backbone. Different VLMs were evaluated in the following subsection. Table~\ref{tab:ablacion} shows the ablation study. In the first four ablation settings (rows 1-4), no explicit past-event description is used. In these first cases, the Reconstruction Prompt includes only its base component (highlighted in blue, Section 3.3), adding the following instruction:
\texttt{``taking into account the attached <RGB/Thermal/UV> images.''} 

\begin{table}[t] 
  \centering
  \caption{Ablation Study with Different Input Modality Configurations ($\Delta = 30$s)}
  \label{tab:ablacion}
  \setlength{\tabcolsep}{2.0pt} 
  \renewcommand{\arraystretch}{1.2} 
  
  \resizebox{\columnwidth}{!}{ 
    \begin{tabular}{cccc|cc|cc|cc}
      \toprule
      \multicolumn{3}{c}{\textbf{Sensors}} & \textbf{Descript.} & \multicolumn{2}{c|}{\textbf{Low-Level}} & \multicolumn{2}{c|}{\textbf{Perceptual}} & \multicolumn{2}{c}{\textbf{High-Level}} \\
      \textcolor{red}{R}\textcolor{mygreen}{G}\textcolor{blue}{B} & \textcolor{red}{TH} & \textbf{UV} & \textbf{SPED} & \textbf{PSNR}$\uparrow$ & \textbf{SSIM}$\uparrow$ & \textbf{LPIPS}$\downarrow$ & \textbf{CLIP}$\uparrow$ & \textbf{AO}$\uparrow$ & \textbf{IoU}$\uparrow$ \\
      \midrule
      
      \cmark & \xmark & \xmark & \xmark & 11.14 & 0.477 & \underline{0.446} & 62.26 & 0.617 & 0.413 \\
      \cmark & \cmark & \xmark & \xmark & 14.49 & 0.558 & 0.537 & 77.55 & \underline{0.818} & 0.567 \\
      \cmark & \xmark & \cmark & \xmark & 13.79 & 0.505 & 0.589 & 76.93 & 0.791 & \underline{0.638} \\
      \cmark & \cmark & \cmark & \xmark & \underline{14.88} & \underline{0.586} & 0.603 & \underline{78.47} & 0.748 & 0.570 \\
      \cmark & \cmark & \cmark & \cmark & \textbf{18.27} & \textbf{0.718} & \textbf{0.297} & \textbf{92.76} & \textbf{0.854} & \textbf{0.739} \\
      
      \bottomrule
    \end{tabular}
  }
\end{table}

Quantitative results are summarized in Table~\ref{tab:ablacion}. Several observations emerge. First, RGB-only inputs produce relatively poor performance across all metrics, which is expected since no residual physical evidence is available. When a single additional modality (TH or UV) is incorporated, performance improves by approximately 2 dB in PSNR, around 10 points in CLIP score, and up to 20\% in AO. We also observe high variance between RGB+TH and RGB+UV configurations, which is reduced when all three modalities are used, suggesting that different traces are informative depending on the interaction type. For instance, thermal cues are particularly informative for sitting interactions, while UV signals are more indicative of touch-based or residual-based events. The largest improvement is obtained when semantic descriptions are incorporated, as the model first reasons about the scene before reconstruction. In this setting, the CLIP score reaches 92.76, indicating strong semantic alignment. Interestingly, low-level metrics such as PSNR remain relatively low ($<$20), reflecting the sensitivity of pixel-level measures to viewpoint variations, while feature-level metrics provide a more reliable indication of reconstruction quality. High-level metrics, which focus on identifying interacting objects and their spatial relationships, confirm that these structural properties are largely preserved, showing significant improvement compared to using RGB-only. Overall, this ablation study demonstrates that multimodal evidence and structured semantic descriptions are both critical for constraining the reconstruction of the past scene.

\begin{figure}[b]
    \centering 
    \includegraphics[width=1\linewidth]{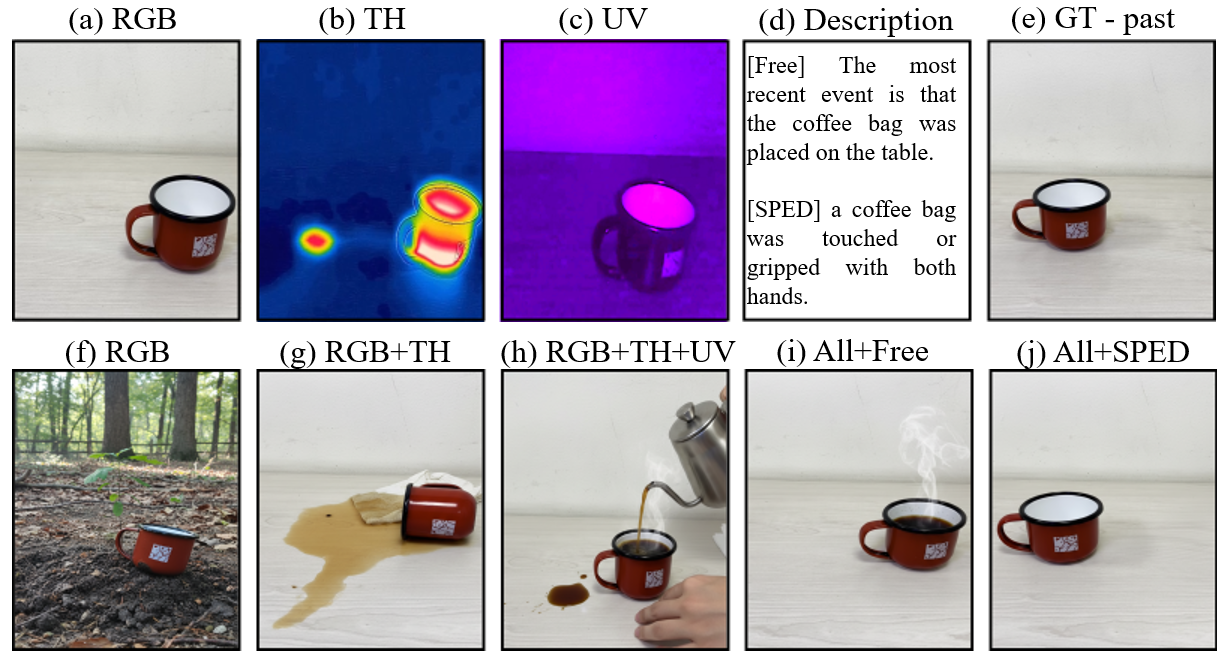}
    \caption{
\textbf{Visual Results of the Ablation Study.}
(a–c) Input RGB, thermal (TH), and ultraviolet (UV) modalities, respectively.
(d) Predicted structured description $\hat{y}_{t-\Delta}$ obtained using both Free and SPED prompts.
(e) Ground-truth past RGB frame.
(f–j) Reconstructed past events under different ablation configurations, highlighting the contribution of each modality and prompting strategy.
}
    \label{fig:cross_domain_visual}
\end{figure}

Figure~\ref{fig:cross_domain_visual} presents a visual comparison of the experiment. Panels (a–c) show the model inputs, (d) illustrates the two description strategies used to constrain the generator, (e) corresponds to the ground truth (GT), and (f–j) display the reconstructed results. RGB-only inputs fail to recover the event and modify the scene without contextual guidance, while incorporating thermal data partially restores the interaction region. Adding UV cues improves finer object details and surface reflections, and integrating descriptive prompts further constrains the reconstruction toward physically plausible outcomes. When guided by the semantic inference module All(RGB+UV+TH)+SPED, the reconstructed scene achieves the highest temporal fidelity, closely matching the ground-truth frame.

\subsection{VLM-Guided Generator Models}
The performance of the generator directly determines the perceptual quality and accuracy of past-scene reconstructions. Beyond predicting the correct past configuration, the model must render a natural and physically consistent scene aligned with the current RGB observation while interpreting thermal and UV traces as evidence of prior interactions. To evaluate this capability, we fix a backward time of $\Delta=60s$ to assess a challenging scenario.

\begin{table}[!t] 
\centering
\caption{Comparison of Generative Models for Time-Reversed Imaging (60\,s Delay). Best Performance in Bold; Second-Best Underlined.} 
\label{tab:comparativa_metodos}
\setlength{\tabcolsep}{4.0pt} 
\renewcommand{\arraystretch}{1.2} 
\resizebox{\columnwidth}{!}{
\begin{tabular}{l|cc|cc|cc}
\toprule
\textbf{Method}
& \multicolumn{2}{c|}{\textbf{Low-Level}}
& \multicolumn{2}{c|}{\textbf{Feature-Level}}
& \multicolumn{2}{c}{\textbf{High-Level}} \\
\cmidrule(lr){2-3}\cmidrule(lr){4-5}\cmidrule(lr){6-7}
& PSNR $\uparrow$ & SSIM $\uparrow$ & LPIPS $\downarrow$ & CLIP $\uparrow$ & AO $\uparrow$ & IoU $\uparrow$ \\
\midrule
Flux Kontext 4.0 & 13.78 & 0.512 & 0.508 & 90.19 & 0.589 & 0.546 \\
Gemini 2.5       & \textbf{16.00} & \textbf{0.615} & \textbf{0.335} & \underline{92.66} & \textbf{0.912} & \textbf{0.851} \\
Dalle 3          & 13.75 & 0.515 & 0.417 & 90.85 & 0.825 & 0.785 \\
Grok             & 13.39 & 0.575 & 0.511 & 84.72 & 0.633 & 0.544 \\
Qwen-Image       & 14.24 & 0.448 & 0.462 & 80.30 & 0.831 & 0.800 \\
SeedDream 4.0    & \underline{15.74} & \underline{0.538} & \underline{0.338} & \textbf{92.88} & \underline{0.879} & \underline{0.830} \\
\bottomrule
\end{tabular}}
\end{table}

We evaluate several VLM-guided generators under the same configuration (All + SPED), including DALL·E~3~\cite{openai2023dalle3}, Grok~\cite{xai2025grok}, Qwen-Image~\cite{pixverse2025}, Flux Kontext~4.0~\cite{batifol2025flux}, Seeddream~4.0~\cite{seedream2025seedream}, and Gemini 2.5~\cite{google2025gemini2_5}. Table~\ref{tab:comparativa_metodos} reports the results. Low-level metrics yield low scores, as they strongly penalize minor variations in viewpoint, pose, or object placement, whereas feature-level metrics are less sensitive to such differences. Gemini 2.5 and Seeddream show more consistent performance. Notably, under high-level evaluation, Gemini 2.5 achieves the strongest results, preserving object identity and spatial layout.

\begin{figure}[!b]
    \centering
    \includegraphics[width=\linewidth]{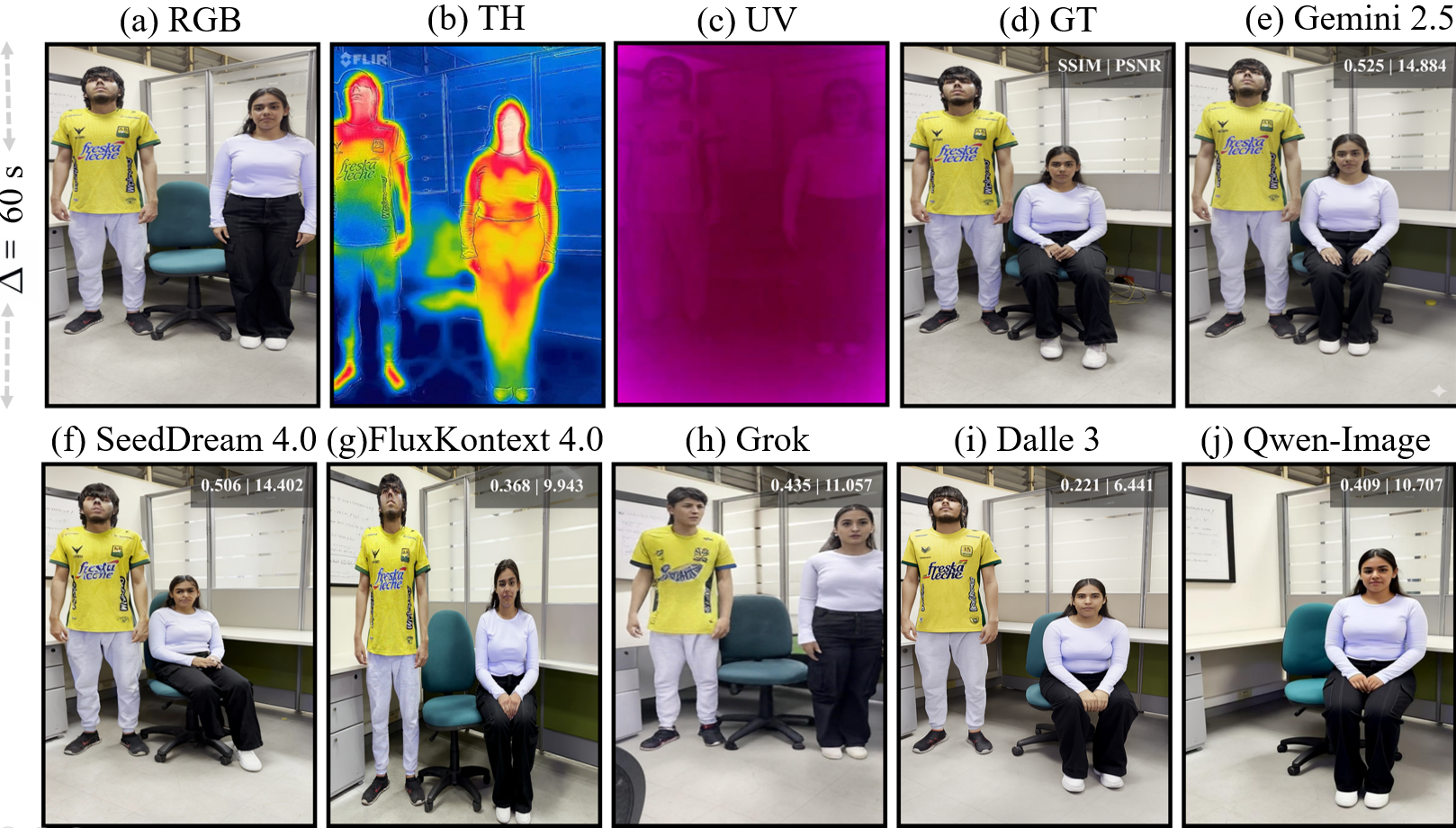}
    \caption{\textbf{Comparison of VLM-guided generators for time-reversed scene reconstruction.}
Each model receives the same multimodal inputs (RGB, Thermal, UV) and prompt.
Gemini 2.5 achieves the closest match to the ground truth (GT), 
maintaining spatial structure and correctly interpreting multimodal trace cues.
}
    \label{fig:generator_models} 
\end{figure}

Figure~\ref{fig:generator_models} illustrates these differences. Grok alters facial identity and introduces inconsistencies in human figures. Qwen-Image occasionally removes secondary subjects, producing incomplete reconstructions. Flux Kontext~4.0 preserves structure but misplaces the seated subject, violating physical constraints. Seeddream~4.0 captures scene context but changes perspective and applies a zoom-out effect, slightly degrading quantitative metrics. DALL·E~3 generates visually coherent images but oversmooths regions and reorganizes elements. In contrast, Gemini 2.5 produces reconstructions that most closely match the GT, accurately preserving spatial layout and interaction semantics. Additional qualitative results in the SM further support these findings, showing that Gemini 2.5 consistently provides the most reliable and temporally coherent reconstructions.

\subsection{Temporal Reconstruction Range} 

This experiment evaluates how far into the past a scene can be reliably reconstructed from residual traces. 
For this analysis, we generated reconstructions at increasing time delays 
($5$\,s, $15$\,s, $30$\,s, $60$\,s, $120$\,s, and $180$\,s) and compared each output 
to the GT. 
All reconstructions were obtained using the complete configuration 
(All + SPED) and the same generator model Gemini 2.5.

\begin{table}[!t] 
\centering
\caption{Temporal Reconstruction Range. Performance as Reconstruction Delay Increases, Reflecting the Progressive Decay of Residual Traces.} 
\label{tab:temporal_range}
\setlength{\tabcolsep}{4.5pt} 
\renewcommand{\arraystretch}{1.2} 
\resizebox{\columnwidth}{!}{
\begin{tabular}{c|cc|cc|cc}
\toprule
\textbf{Delay (s)}
& \multicolumn{2}{c|}{\textbf{Low-Level}}
& \multicolumn{2}{c|}{\textbf{Perceptual}}
& \multicolumn{2}{c}{\textbf{High-Level}} \\
\cmidrule(lr){2-3}\cmidrule(lr){4-5}\cmidrule(lr){6-7}
& PSNR $\uparrow$ & SSIM $\uparrow$ & LPIPS $\downarrow$ & CLIP $\uparrow$ & AO $\uparrow$ & IoU $\uparrow$ \\
\midrule
5   & 18.55 & 0.776 & 0.325 & 91.93 & 0.919 & 0.887 \\
15  & 18.55 & 0.775 & 0.321 & 91.98 & 0.918 & 0.873 \\
30  & 18.27 & 0.718 & 0.297 & 92.76 & 0.854 & 0.739 \\
\rowcolor{gray!15} 60  & 16.00 & 0.615 & 0.335 & 92.66 & 0.912 & 0.851 \\
\rowcolor{gray!25} 120 & 14.67 & 0.631 & 0.325 & 86.72 & 0.774 & 0.711 \\
\rowcolor{gray!35} 180 & 12.14 & 0.522 & 0.213 & 75.14 & 0.643 & 0.511 \\
\bottomrule
\end{tabular}}
\end{table}

Table~\ref{tab:temporal_range} reports quantitative results across different time delays. Performance remains relatively stable up to $30s$, degrades at $60s$, and becomes significantly worse at $120s$, where reconstructions lose accuracy. High-level metrics (AO and IoU) remain relatively robust up to $60s$, indicating that object identity and spatial relationships are still preserved within this temporal range. Beyond this point, the physical evidence captured by thermal and UV modalities fades, increasing uncertainty in both reconstruction accuracy and semantic consistency.

\begin{figure}[t]
\centering
\includegraphics[width=\linewidth]{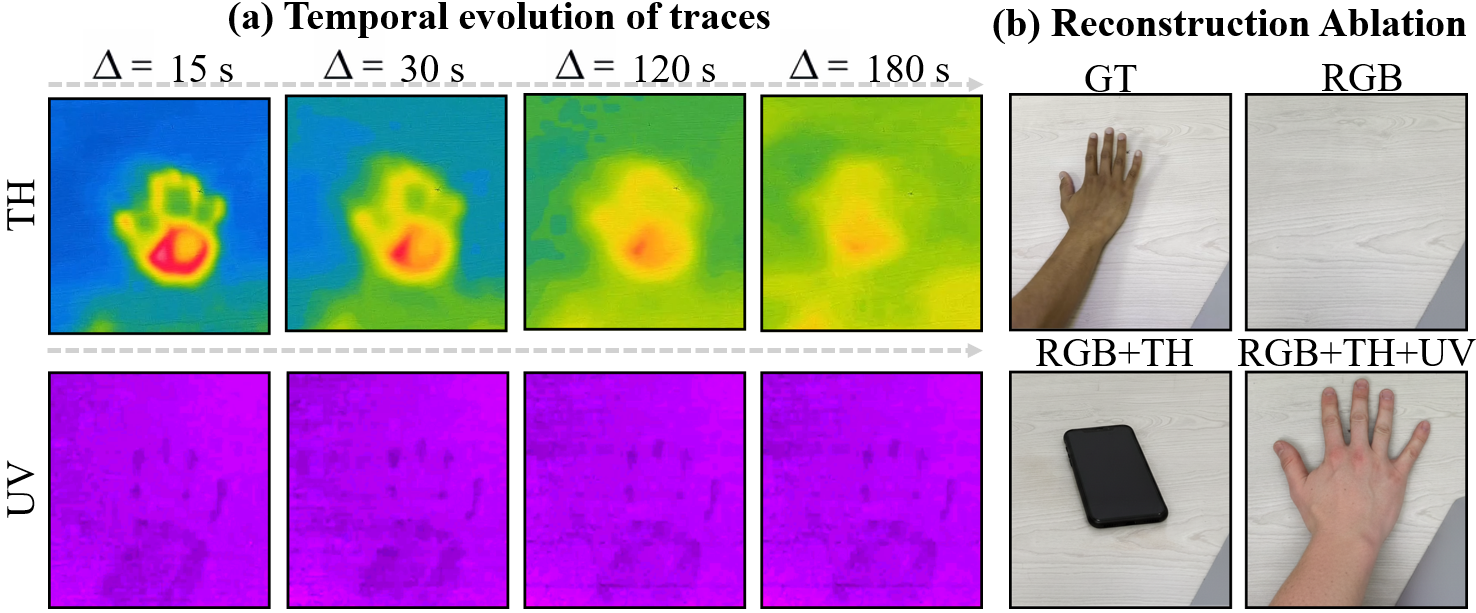}
\caption{
(\textbf{a}) Temporal evolution of TH and UV traces. (\textbf{b}) Reconstruction ablation: RGB-only fails to infer the event, RGB+TH misinterprets weak thermal cues, while RGB+TH+UV produces reconstructions closer to the GT.}
\label{fig:temporal_decay}
\end{figure}

Figure~\ref{fig:temporal_decay}(\textbf{a}) illustrates how traces evolve over time. Immediately after interaction, thermal signals clearly preserve the hand structure but gradually dissipate, becoming nearly imperceptible after $120s$. In contrast, UV traces persist longer and remain partially visible at later times. This behavior explains the higher variance observed at $180s$ in Table~\ref{tab:temporal_range}, where some scenes benefit from persistent UV fluorescence while others suffer from the disappearance of thermal cues. Figure~\ref{fig:temporal_decay}(\textbf{b}) shows reconstruction ablations at $180s$. With RGB-only input, the model lacks evidence of prior interaction and simply preserves the current scene. RGB+TH inputs provide weak constraints at long delays, often leading to ambiguous or hallucinated reconstructions.  UV traces, however, retain the hand outline and offer stronger guidance. When all modalities are combined, the system exploits both early thermal decay and persistent UV cues, producing reconstructions closer to the ground truth. These results highlight how multimodal traces progressively constrain the set of plausible past states.

\subsection{Robustness to Physics-Informed Time Estimation}

The proposed framework assumes that the elapsed time $\Delta$ between the observed thermal image and the contact event is available as an input. To evaluate the sensitivity of the method to inaccuracies in this parameter, we consider a simple physics-informed estimator based on the thermal cooling process. Assuming that residual thermal traces approximately follow Newton's law of cooling, the temperature evolution is modeled as

\begin{equation}
T(t)=T_{\mathrm{env}}+\left(T_0-T_{\mathrm{env}}\right)e^{-kt},
\end{equation}

where $T_{\mathrm{env}}=28^\circ$C denotes the ambient temperature, $T_0=34^\circ$C is the initial contact temperature, and $k=0.03\,\mathrm{s}^{-1}$ is a constant cooling coefficient representing typical indoor conditions. The elapsed time is estimated by inverting the model,

\begin{equation}
\hat{t}=-\frac{1}{k}\log\left(\frac{T(t)-T_{\mathrm{env}}}{T_0-T_{\mathrm{env}}}\right).
\end{equation}

In practice, the residual contact region is obtained by thresholding pixels above the ambient temperature, and the average temperature within this region is used to estimate $\hat{t}$. The estimated delay is then used as the temporal input of the proposed reverse-time inference framework.

Table~\ref{tab:physics_delta} reports the reconstruction performance for three representative reconstruction delays (30\,s, 60\,s, and 120\,s), comparing the use of the exact elapsed time with the physics-informed estimate. The thermal model predicts delays of approximately $27\pm2.4$\,s, $56\pm2.8$\,s, and $114\pm3.5$\,s for the three settings, respectively. Across all reconstruction delays, replacing the ground-truth temporal input with the estimated one leads to only a modest reduction in reconstruction quality. The largest degradation is observed at the shortest delay, where a small temporal error represents a larger fraction of the elapsed time, while the performance at longer delays remains comparatively stable. High-level semantic consistency (CLIP-I) is largely preserved, and the action localization metrics (AO and IoU) exhibit only minor decreases. These results suggest that the proposed framework is robust to moderate errors in the estimated action time, indicating that simple physics-based thermal models can provide sufficiently accurate temporal initialization when the exact elapsed time is unavailable.

\begin{table}[t]
\centering
\small
\caption{Robustness to physics-informed temporal estimation.}
\label{tab:physics_delta}
\renewcommand{\arraystretch}{1.2}
\resizebox{0.98\columnwidth}{!}{%
\begin{tabular}{c|c|cc|cc|cc}
\toprule
\textbf{Delay} & \textbf{Time Input} & \multicolumn{2}{c|}{\textbf{Low-Level}} & \multicolumn{2}{c|}{\textbf{Feature-Level}} & \multicolumn{2}{c}{\textbf{High-Level}} \\
\cmidrule(lr){3-4} \cmidrule(lr){5-6} \cmidrule(lr){7-8}
 & & \textbf{PSNR $\uparrow$} & \textbf{SSIM $\uparrow$} & \textbf{LPIPS $\downarrow$} & \textbf{CLIP-I $\uparrow$} & \textbf{AO $\uparrow$} & \textbf{IoU $\uparrow$} \\
\midrule

\rowcolor{green!15}\multirow{2}{*}{30s} & Exact, $\Delta=30$s & 18.268 & 0.718 & 0.297 & 92.755 & 0.854 & 0.739 \\
 & Estimated, $\hat{\Delta}=27$s & 17.421 & 0.691 & 0.315 & 92.884 & 0.811 & 0.701 \\
\midrule

\rowcolor{green!15}\multirow{2}{*}{60s} & Exact, $\Delta=60$s & 16.000 & 0.615 & 0.335 & 92.660 & 0.912 & 0.851 \\
 & Estimated, $\hat{\Delta}=56$s & 15.563 & 0.603 & 0.344 & 92.731 & 0.891 & 0.829 \\
\midrule

\rowcolor{green!15}\multirow{2}{*}{120s} & Exact, $\Delta=120$s & 14.670 & 0.631 & 0.325 & 86.720 & 0.774 & 0.711 \\
 & Estimated, $\hat{\Delta}=114$s & 14.382 & 0.621 & 0.332 & 86.804 & 0.758 & 0.694 \\
\bottomrule
\end{tabular}%
}
\vspace{-1em}
\end{table}

\section{Conclusion and Discussion} 

This work introduced Time-Reversed Imaging, a new paradigm in computer vision that seeks to reconstruct \textit{what just happened} by leveraging multimodal residual traces left in the environment. Through the TRACE-HEI dataset, we established the first benchmark explicitly designed to study residual trace dynamics and past-event inference. Our results show that thermal and UV signals encode complementary temporal evidence that, when fused with RGB observations and structured semantic priors, enables physically grounded reconstruction of recent scene states. The proposed framework, combining multimodal sensing, semantic reasoning, and VLM-guided diffusion, moves beyond statistical extrapolation and toward inference constrained by residual physics.

Empirically, we observe that thermal diffusion and fluorescence persistence provide reliable temporal constraints up to meaningful time horizons, reducing ambiguity in reconstruction and improving semantic consistency. While low-level similarity degrades as traces fade, high-level object and interaction fidelity remain robust when guided by multimodal evidence. These findings suggest that perception can be extended beyond the visible present: cameras need not only observe the current state but can computationally reason about its recent causes.

Time-Reversed Imaging opens exciting research avenues. Looking forward, extending temporal horizons will require explicit modeling of residual trace dynamics as continuous decay processes governed by material-dependent heat diffusion and photoluminescence. Incorporating additional sensing modalities could further constrain the inverse problem. Expanding to outdoor environments is feasible but introduces challenges including uncontrolled illumination, wind-driven thermal dissipation, and UV variability, necessitating adaptive normalization, environmental calibration, and physics-aware modeling. By addressing these challenges, Time-Reversed Imaging can evolve into a broader framework for temporal reasoning, bridging multimodal sensing, inverse problems, and generative modeling to reconstruct the recent past rather than merely predict the future.

Finally, the capacity to computationally recover unrecorded past states necessitates addressing critical societal implications. While this paradigm offers substantial benefits for forensic analysis, human-robot collaboration, and safety monitoring, it inherently introduces distinct privacy and surveillance risks. The ability to deduce prior human actions from residual environmental clues must be handled responsibly. Consequently, the future development and deployment of time-reversed inference systems must be systematically aligned with robust ethical guidelines, privacy-preserving algorithms, and rigorous regulatory frameworks

\bibliographystyle{IEEEtran}
\bibliography{main}

\end{document}